\documentclass[conference, a4paper]{IEEEtran}
\IEEEoverridecommandlockouts
\usepackage{booktabs}
\usepackage{cite}
\usepackage{amsmath,amssymb,amsfonts}
\usepackage{algorithmic}
\usepackage{graphicx}
\usepackage{textcomp}
\usepackage{xcolor}
\usepackage{balance}
\def\BibTeX{{\rm B\kern-.05em{\sc i\kern-.025em b}\kern-.08em
    T\kern-.1667em\lower.7ex\hbox{E}\kern-.125emX}}
\begin{document}

% \title{Segmenting messy text: detecting boundaries of topically similar segments in text derived from images of historical newspapers\\
% }

\title{Segmenting Messy Text: Detecting Boundaries in Text Derived from Historical Newspaper Images\\
}

\author{\IEEEauthorblockN{1\textsuperscript{st} Carol Anderson}
\IEEEauthorblockA{
\textit{Ancestry.com}\\
Lehi, Utah, USA \\
caanderson@ancestry.com}
\and
\IEEEauthorblockN{2\textsuperscript{nd} Phil Crone}
\IEEEauthorblockA{
\textit{Ancestry.com}\\
San Francisco, California, USA\\
pcrone@ancestry.com}
}

\maketitle

\begin{abstract}
Text segmentation, the task of dividing a document into sections, is often a prerequisite for performing additional natural language processing tasks. Existing text segmentation methods have typically been developed and tested using clean, narrative-style text with segments containing distinct topics. Here we consider a challenging text segmentation task: dividing newspaper marriage announcement lists into units of one announcement each. In many cases the information is not structured into sentences, and adjacent segments are not topically distinct from each other. In addition, the text of the announcements, which is derived from images of historical newspapers via optical character recognition, contains many typographical errors. As a result, these announcements are not amenable to segmentation with existing techniques. We present a novel deep learning-based model for segmenting such text and show that it significantly outperforms an existing state-of-the-art method on our task. 
\end{abstract}


\section{Introduction}
Text segmentation is the task of dividing a document into sections, with the granularity of segmentation varying depending on the application. For example, both the segmentation of news feeds into topically distinct articles \cite{beeferman1999statistical} and the segmentation of character sequences into words \cite{webster-1992} can be considered forms of text segmentation. Text segmentation facilitates many downstream natural language processing tasks, including information extraction, text summarization, and passage retrieval. Here, we focus on a level of segmentation that we will refer to as {\sc Topic Segmentation}: dividing a document into sections with topically distinct content. We use the term {\sc Topic Segmentation} to distinguish this level of text segmentation from more granular types of segmentation, such as word or sentence segmentation.\footnote{This terminology may be less than ideal for the case we describe, since, as we note below, the segments in our data could be seen as sharing a single topic (marriage). At the same time, these segments describe distinct events (weddings of different couples). For this reason, we take {\sc Topic Segmentation} to be an appropriate description of our task.}

Much of the published research on topic segmentation has focused on segmenting clean blocks of narrative-style text, such as news articles or Wikipedia pages. Traditional approaches to these segmentation tasks detect boundaries between topics by unsupervised methods, for example by measuring lexical cohesion \cite{hearst-1997} or by explicitly modeling topics with methods such as latent Dirichlet allocation (LDA) \cite{riedl2012topictiling}. In recent years, supervised approaches have been shown to be highly successful at detecting transitions between topics. The current state-of-the-art topic segmentation methods use deep neural networks to predict whether a given sentence marks the boundary of a segment \cite{koshorek-etal-2018-text,chen2009global,glavavs2020two}.

\begin{figure}
    \centering
    \includegraphics{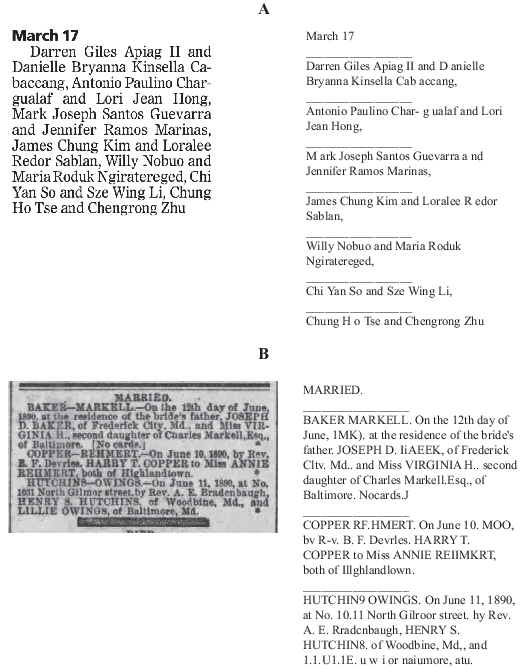}
    \caption{Examples of articles in our data set shown along with properly segmented extracted text. A) Article from the \textit{Pacific Daily News}, March 19, 2016 (p. A26), www.newspapers.com/clip/20749217. Text was extracted from a PDF.  B) Article from \textit{The Baltimore Sun}, June 13, 1890 (p. 2), www.newspapers.com/clip/23188935. Text was extracted with ABBYY FineReader 9.} 
    \label{fig:seg_examples}
\end{figure}

\begin{figure}
    \centering
    \includegraphics{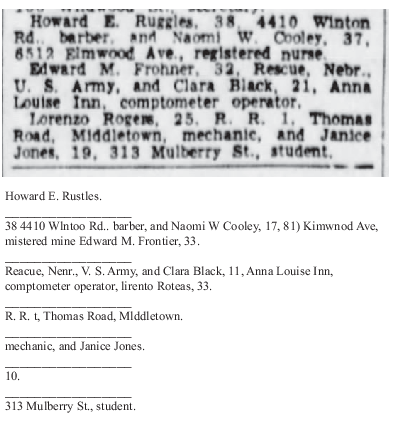}
    \caption{Sentence splitting performs poorly on newspaper marriage announcements. The output of the NLTK Punkt sentence tokenizer is shown. Article was published in \textit{The Cincinnati Enquirer} on June 22, 1954 (p. 11), www.newspapers.com/clip/9845381. Due to space constraints, only part of the article is shown here.}
    \label{fig:sentence_tokenization}
\end{figure}

Here we consider a topic segmentation task not well solved by current methods. Our goal is to segment the text of newspaper marriage announcement listings into units of one announcement each (Fig. \ref{fig:seg_examples}). The segmentation task is part of a larger pipeline to extract genealogical information from images of historical newspapers. Our task differs from previous work on topic segmentation discussed above in the following ways:
\begin{enumerate}
    \item {\sc Text derived from images} \quad The text of the newspaper pages has been extracted from images by optical character recognition (OCR) software or from PDF documents. Many of the newspapers were yellowed or faded prior to scanning, while others were transferred from relatively low resolution microfilm images. As a result, this text contains errors such as wrong letters, mis-capitalization, and punctuation mistakes. Even extraction from PDFs can yield imperfect text (Fig. \ref{fig:seg_examples}A). This makes sentence segmentation, upon which many existing topic segmentation methods depend, difficult (Fig. \ref{fig:sentence_tokenization}). Cues normally indicating sentence transitions, such as capitalization of the first word in a sentence or a period at the end of a sentence, may be missing or spurious. On the other hand, OCR and PDF extraction software produces information about the spatial layout of the text, which can aid in segmentation. 
    \item {\sc Non-narrative structure} \quad Newspaper marriage lists do not typically contain full sentences. These lists usually consist of a series of names, locations, and other facts. This again makes sentence segmentation difficult.
    \item {\sc Topical similarity} \quad Segments describe different marriage events involving different couples, but tend to be lexically and semantically similar to each other. Many existing topic segmentation methods rely on detecting adjacent, but lexically or semantically dissimilar sections of text, which is impractical in our case.
    \item {\sc Hierarchical structure} \quad The newspaper articles may contain internal subheadings, such as dates or places, that apply to multiple couples. A strictly linear segmentation approach, in which only the boundaries between segments are predicted, would not fully capture the underlying structure of the document. 
\end{enumerate}

While we focus here on the case of marriage announcements, historical newspapers contain many similar types of listings that convey other forms of information, e.g. death lists, birth lists, lists of real estate transactions, etc. The challenges and methods described here are relevant to work on segmenting these listings, as well as segmenting other types of noisy text, such as text extracted automatically from financial prospectuses \cite{AzziTheFS} or from voice-to-text transcription \cite{sheikh2017topic}.

Our proposal differs from much of the existing work on topic segmentation in several key respects: 
\begin{enumerate}
    \item Since sentence splitting could not be done faithfully with our text, we predict boundaries at the token level, rather than at the sentence level. 
    \item Because our text is derived from images, and OCR or PDF extraction software provides the physical location of each token, we use the locations of tokens as a feature of our model. This has the potential to aid the detection of the beginnings of segments, since they often start at the beginning of a new line of text. 
    \item We use ELMo embeddings \cite{peters-etal-2018-deep} as a feature of our model. We fine-tune the language model from which the ELMo embeddings are generated on a large corpus of newspaper text extracted from images. This fine-tuning allows the ELMo model to generate both newspaper-specific embeddings and embeddings that capture the meanings of words with common OCR errors.
    \item Given the hierarchy of information within our text, we do not approach the task as strictly linear segmentation. Rather than just predicting boundaries between segments, we predict whether each token is the beginning of a segment (\texttt{B-Marriage}), inside a segment (\texttt{I-Marriage}), or outside a segment (\texttt{O}). In practice, we use this segmentation system in conjunction with another sequence-tagging model that we trained to label key marriage facts, such as \texttt{Bride}, \texttt{Groom}, \texttt{WeddingDate}, etc. We assume that information falling outside a segment, such as a \texttt{WeddingDate}, applies to all couples after it, until another \texttt{WeddingDate} is reached or the article ends. 
\end{enumerate}
As a point of comparison, we show that on our data set, the model described here significantly outperforms a topic segmentation model recently proposed by Koshorek et al.~\cite{koshorek-etal-2018-text}.

\section{Related work}
Two types of approaches have dominated previous work on topic segmentation. The first approach is unsupervised and attempts to determine the lexical, semantic, or topical similarity between adjacent sections of text. Contiguous sections that are highly similar are taken to constitute a segment, while segment boundaries are detected by finding adjacent sections of text that are dissimilar. The second approach uses supervised machine learning methods that are trained on data labeled with segment boundaries. In some cases, these supervised models also leverage the fact that segments should be topically similar to solve the problem of identifying segment boundaries.

% An early example of the first approach is offered by Hearst \cite{hearst1994multi}. Hearst's method utilizes a bag-of-words model that represents sections of text with TF-IDF vectors. Cosine similarities between the vectors of adjacent sections are then computed, and segment boundaries are inferred to exist where large decreases in these similarities are detected. Choi \cite{choi2000advances} also makes use of a bag-of-words model, but computes the similarity between all sentence pairs in the document. A clustering algorithm is then used to group together adjacent sentences into segments. Riedl and Biemann \cite{riedl2012topictiling} introduce a model similar to Hearst's, but which uses LDA \cite{blei2003latent} to assign topics to each word in a document. Riedl and Biemann represent sections of text as bags of topics, rather than bags of words, and use the resulting topic vectors to compute the similarity between adjacent sections of text.

One of the earliest unsupervised topic segmentation models is TextTiling \cite{hearst1994multi}, in which term frequencies are used to measure the similarity between adjacent sections of text. Boundaries are inferred to exist between highly dissimilar sections. Choi \cite{choi2000advances} also uses term frequencies, but computes the similarity between all sentence pairs in the document and uses a clustering algorithm to group together adjacent sentences into segments. Riedl and Biemann \cite{riedl2012topictiling} extend TextTiling \cite{hearst1994multi} by using LDA \cite{blei2003latent} to assign topics to each word in a document. Chen et al. \cite{chen2009global} introduce a generative model that, like LDA, treats documents as a bag of topics. However, unlike LDA, \cite{chen2009global} assumes that each sentence is assigned to a single topic, that all sentences sharing the same topic are adjacent, and that topics follow a predictable order. In contrast, Mota et al.'s BeamSeg model \cite{mota-etal-2019-beamseg} assumes that multiple segments within a document may share the same topic. During inference, BeamSeg uses a beam search algorithm to find an optimal segmentation. Glava\v{s} et al.~\cite{glavavs2016unsupervised} use word2vec embeddings \cite{mikolov2013efficient} to compute similarity scores between all pairs of sentences in a document. These scores are used to construct a graph of all sentences in the document, and adjacent sentences are then merged into segments via an algorithm that considers the sentences' membership in cliques within the graph.

In contrast to unsupervised approaches, supervised approaches to topic segmentation train machine learning models on data with labeled segment boundaries. For example, Sheikh et al. \cite{sheikh2017topic} represent a document as a sequence of word2vec embeddings, then passes this sequence through a bidirectional LSTM (BiLSTM) \cite{Schuster1997,hochreiter1997long}. Following the intuition that segment boundaries occur where adjacent spans of text are topically dissimilar, \cite{sheikh2017topic} uses the cosine distance between the output from the forward and backward LSTMs at each position to predict the presence of a segment boundary at that position. Arnold et al. \cite{arnold_et_al_2019} use a BiLSTM-based approach in which a probability distribution over topics is produced for each sentence in a document. Segment boundaries are predicted where adjacent sentences have dissimilar predicted topic distributions.

Other supervised approaches \cite{koshorek-etal-2018-text,chang2019language,glavavs2020two} make use of a hierarchical structure in which sentence embeddings are learned from pre-trained token embeddings. The resulting sequence of sentence embeddings is then passed through a sequence-to-sequence layer, e.g. a recurrent neural network (RNN), a self-attention layer, etc., and each element of the output sequence is passed to a final output layer. Koshorek et al.~\cite{koshorek-etal-2018-text} use such an approach, based on BiLSTMs, to learn sentence representations from pre-trained word2vec embeddings and then to classify each sentence as either the last in a segment, or not the last in a segment. Chang et al.~\cite{chang2019language} use similar architecture, but incorporate language model pre-training into their model. Glava\v{s} and Somasundaran~\cite{glavavs2020two} also learn sentence level embeddings from token level embeddings, but use self-attention based encoder stacks \cite{vaswani2017attention} rather than RNNs.

Most recent work on topic segmentation follows the supervised approach, and supervised models of topic segmentation have achieved state-of-the-art results on public data sets in recent years \cite{koshorek-etal-2018-text,chang2019language,glavavs2020two}. We likewise follow the supervised approach. We do so not only because of the demonstrated success of supervised approaches, but because, as noted above, the segments of the newspaper articles whose boundaries we attempt to detect are highly similar in lexical and semantic content. This makes it unlikely that an unsupervised approach attempting to detect topical discontinuities between adjacent sections of text can be fruitfully applied to our use case.

While the most salient distinction between previous approaches to topic segmentation is that between supervised and unsupervised approaches, previous approaches can also be distinguished with respect to whether they attempt only to segment text \cite{hearst1994multi,choi2000advances,riedl2012topictiling,glavavs2016unsupervised,sheikh2017topic,koshorek-etal-2018-text,chang2019language,glavavs2020two} or also attempt to classify the topic or type of each section \cite{chen2009global,arnold_et_al_2019,mota-etal-2019-beamseg}. While we focus only on attempting to detect one topic, marriage announcement, our approach could easily be extended to a larger number of topics.

\begin{figure*}
    \centering
    \includegraphics{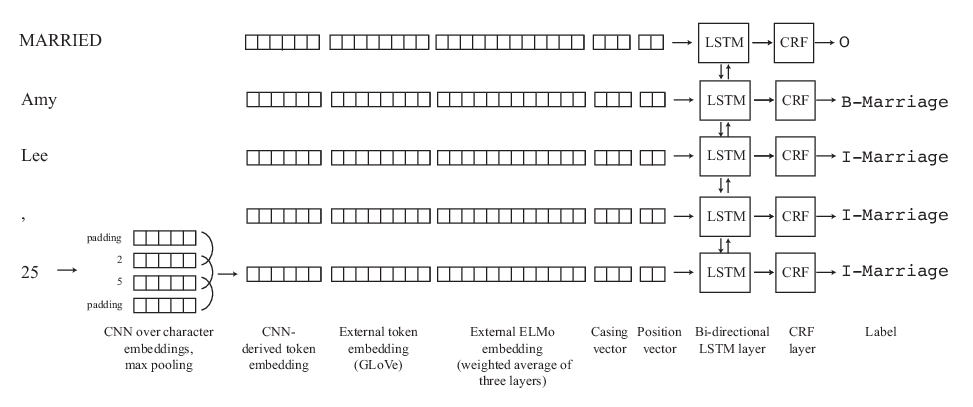}
    \caption{Illustration of our model architecture. Vectors are not shown to scale. The first token in this example, \textit{MARRIED}, is not part of a specific marriage announcement and is therefore labeled \texttt{O}. The first segment begins with \textit{Amy}.}
    \label{fig:arch}
\end{figure*}

In general, previous work on topic segmentation has made use of resources such as Wikipedia, which contain segment information in the form of section breaks \cite{chen2009global,koshorek-etal-2018-text} or has constructed synthetic data sets by concatenating the text of different news articles \cite{choi2000advances,sheikh2017topic}. The closest analogue to the data set used in the current work is a corpus of newspaper ads used by Manning in \cite{Manning_rethinkingtext}. 
The lack of complete sentences, topical similarity of different segments, and hierarchy of information are all found in \cite{Manning_rethinkingtext} as well. However, to the best of our knowledge, no prior work has attempted segmentation on text derived from newspaper images pre-processed with OCR software. Our work also differs from most previous work on topic segmentation in that we attempt to segment documents at the token level, rather than the sentence level. To the best of our knowledge, among the recent approaches to topic segmentation, only \cite{sheikh2017topic} attempts to segment documents at the token level. As mentioned above, the nature of our data makes it difficult to successfully pre-process text by segmenting it into sentences or paragraphs. For this reason, detecting segment boundaries at a level of less granularity than the token level is infeasible. In this regard, our formulation of the problem is similar to the task of sentence boundary detection in noisy text \cite{AzziTheFS}.

\section{Data \& Model}
\subsection{Data}
Our entire data set consists of 1384 newspaper articles. The text of each newspaper page was extracted using the following tools: ABBYY FineReader 9 (1169 articles), the commercial OCR packages ABBYY FineReader 11 and OmniPage 15 (10 articles), and the PDF text extractors iText or Apache PDFBox (205 articles). This assortment of different tools is due to changes in our production system over time, as well as changes in newspaper publishing, with more recent content being published as PDFs and older content being digitized by scanning printed pages. Marriage article boundaries were drawn by hand. These articles contain a total of 16,833 segments, with each segment being the description of a particular marriage; the median number of segments per article is 7. The median document length is 610 characters and the median segment length is 63 characters. Segment boundaries were labeled by hand. We divided the data into training, development, and test sets of 804, 303, and 277 articles, respectively.

\subsection{Tagging scheme}
Inspired by the BIO tagging scheme commonly used for named entity recognition \cite{DBLP:journals/corr/cmp-lg-9505040}, we label each token as \texttt{B-Marriage}, \texttt{I-Marriage}, or \texttt{O}, indicating, respectively, that the token marks the beginning of a marriage announcement segment, is inside a segment, or is not in a segment. For example, in Fig. \ref{fig:arch}A, the word \textit{MARRIED} does not belong to a particular segment and would therefore be labeled \texttt{O}. This tagging scheme allows us to simultaneously segment the text and categorize segments. Therefore, our approach can be seen as a form of joint text segmentation and segment classification with one segment class (\texttt{Marriage}). This scheme could be extended to include additional segment classes. In our particular use case, information outside of a marriage announcement often applies to the marriages listed after it, e.g. the location of a county clerk’s office where marriage licenses were granted. The \texttt{O} tags allow us to treat such information differently than we treat segments about a particular couple. 
    
In contrast to our approach, most existing topic segmentation methods formulate the task as a boundary-detection problem: the goal is to find the beginning \textit{or} the end of each segment. It is not necessary to find both the beginnings \textit{and} ends of segments, because the beginning of a new segment is always immediately after the ending of the previous one. 

\subsection{Model architecture}
We frame the task as a token level sequence tagging problem, using an architecture based on those proposed in \cite{lample-etal-2016-neural} and \cite{ma-hovy-2016-end}. Fig. (\ref{fig:arch}) illustrates the overall model architecture.

For each token \(t_i\), the model first computes a character-based representation \(\textbf{t}^{char}_i\). This is accomplished by representing each character in the token as a 25-dimensional learned embedding. Character embeddings are passed through a convolutional layer consisting of 30 filters of kernel size 3, using \(same\) padding, followed by global max pooling. The output of the max pooling layer is concatenated with externally trained token embeddings. For each token, we employ a 100-dimensional GloVe embedding \(\textbf{t}^{glove}_i\) \cite{pennington-etal-2014-glove} and an ELMo embedding generated from the language model described in \cite{peters-etal-2018-deep}. We fine-tune a pre-trained model using the method described in \cite{peters-etal-2018-deep} for 1.3 epochs on a corpus of 1.6 billion tokens derived from newspaper page images by OCR or by extraction from PDF. The ELMo model returns a set of three embeddings for each token. We take a weighted average of these three embeddings, with weights learned during training, to obtain \(\textbf{t}^{elmo}_i\). Since all tokens are lowercased prior to use, we also incorporate an eight-dimensional, one-hot encoded vector \(\textbf{t}^{casing}_i\) to represent the original capitalization of each token (e.g. uppercase, lowercase, etc.), using the method of \cite{JMLR:v12:collobert11a}.

In some experiments we also incorporate a feature to represent the physical locations of tokens. The OCR software produces a bounding box for each token, including the \(x\) and \(y\) pixel coordinates of the top left corner of the token's bounding box. The physical locations are a potential signal for the beginning of a new line of printed text. However, the raw \(x\) and \(y\) coordinates vary dramatically depending on which part of an image an article came from. Therefore, instead of using the raw coordinates, we compute a distance vector \(\textbf{t}^{dist}_i\) between tokens \(t_i\) and \(t_{i-1}\), using the \(x\) and \(y\) coordinates of each. The vector is set to \([0, 0]\) for the first token in an article. 

The full embedding for token \(t_i\), \(\textbf{v}_i\) is given by Equation (\ref{eq:token1}) for experiments that do not make use of distance vectors and by  Equation (\ref{eq:token2}) for experiments that do. In both equations, \(\circ\) denotes concatenation.
\begin{equation}
    \label{eq:token1}
    \textbf{v}_i = \textbf{t}^{char}_i \circ \textbf{t}^{glove}_i \circ \textbf{t}^{elmo}_i \circ \textbf{t}^{casing}_i
\end{equation}
\begin{equation}
    \label{eq:token2}
    \textbf{v}_i = \textbf{t}^{char}_i \circ \textbf{t}^{glove}_i \circ \textbf{t}^{elmo}_i \circ \textbf{t}^{casing}_i \circ \textbf{t}^{dist}_i
\end{equation}

The sequence of all token embeddings for a document of length \(n\), \(\textbf{v}_{1:n}\), is then passed through a single BiLSTM layer, with a state size of 100 for each direction. We apply dropout at a rate of 0.5 prior to feeding token representations to the BiLSTM layer. Let \(\overrightarrow{\textrm{LSTM}}\) and \(\overleftarrow{\textrm{LSTM}}\) denote the forward and backward LSTMs, respectively, and let \(\overrightarrow{\textbf{c}_i}\) and \(\overleftarrow{\textbf{c}_i}\) denote the internal cell states of the forward and backward LSTMs at position \(i\), respectively. Then we obtain a hidden representation \(\textbf{h}_i\) for each token \(t_i\) as follows:
\begin{gather}
    \overrightarrow{\textbf{h}_i}, \overrightarrow{\textbf{c}_i} = \overrightarrow{\textrm{LSTM}}(\textbf{v}_i, \overrightarrow{\textbf{h}_{i-1}}, \overrightarrow{\textbf{c}_{i-1}}) \\
    \overleftarrow{\textbf{h}_i}, \overleftarrow{\textbf{c}_i} = \overleftarrow{\textrm{LSTM}}(\textbf{v}_i, \overleftarrow{\textbf{h}_{i-1}}, \overleftarrow{\textbf{c}_{i - 1}}) \\
    \textbf{h}_i = \overrightarrow{\textbf{h}_i} \circ \overleftarrow{\textbf{h}_i}
\end{gather}
The sequence of hidden outputs from the BiLSTM layer \(\textbf{h}_{1:n}\) are then fed as input to a linear-chain conditional random field (CRF) to produce an output sequence of labels \(\hat{\textbf{y}}_{1:n}\). During inference, the Viterbi algorithm is used to decode the most likely sequence \(\hat{\textbf{y}}_{1:n}\).

All model weights are trained in a supervised manner using the following negative log likelihood loss function \(\mathcal{L}\):
\begin{equation}
    \mathcal{L} = - \sum_k \log p(\textbf{y}_{1:n_k}|\textbf{h}_{1:n_k})
    \label{eq:loss}
\end{equation}
In Equation (\ref{eq:loss}), \(p(\textbf{y}_{1:n_k}|\textbf{h}_{1:n_k})\) denotes the probability assigned by the CRF to true label sequence \(\textbf{y}_{1:n_k}\) for training example \(k\) with \(n_k\) tokens given a sequence of hidden outputs from the BiLSTM layer \(\textbf{h}_{1:n_k}\). We employ minibatch gradient descent with a batch size of 16, using the nadam optimizer \cite{Dozat2016IncorporatingNM}. We use a learning rate of 0.001; other nadam parameters are as described in \cite{Dozat2016IncorporatingNM}. For all other model hyperparameters, we use hyperparameters previously chosen during optimization of a similar model architecture for a different sequence-tagging task using text from the same domain.

\subsection{Evaluation}
\noindent \(\mathbf{P_k}\) \quad For comparison with previous work, we calculate the \(P_k\) metric \cite{beeferman1999statistical}. \(P_k\) is the probability that, when sliding a window of size \textit{k} over predicted segments, the ends of the window are in different segments when they should have been in the same segment, or vice-versa. For calculating \(P_k\), all tokens must be included in a segment beginning with a \texttt{B-Marriage} label. Therefore, prior to computing \(P_k\) we convert any \texttt{O} labels in the predictions and ground truth to \texttt{B-Marriage} or \texttt{I-Marriage} (depending on whether or not the \texttt{O} is the first in a series of \texttt{O}s), so that stretches of \texttt{O} become segments. We use the \texttt{SegEval} package \cite{fournier-2013-evaluating} for calculating \(P_k\) and set \textit{k} to half the average segment size for each document.\\

% Booktabs table:
\begin{table*}[t]
\centering
\caption{Scores for our model and the model proposed in \cite{koshorek-etal-2018-text} trained and tested on marriage announcements. Scores shown are the average of three experiments; standard deviations are given for \(P_k\) and F1. Lower \(P_k\) indicates higher segmentation accuracy.}
\begin{tabular}{@{}llllccc@{}}
\toprule
Model & Features & Labels & \(P_k\) & \multicolumn{3}{c}{Task-Based Evaluation}\\ \cmidrule{5-7}
& & & & Precision & Recall & F1\\ \midrule
Ours & All features & BIO & 0.039 $\pm{0.002}$ & \textbf{97.1}  & 98.6  & \textbf{97.8} $\pm{0.4}$\\
 & ELMo not fine-tuned & BIO & 0.049 $\pm{0.007}$ & 93.3  & 98.1 & 95.6 $\pm{0.7}$\\
 & No ELMo & BIO & 0.078 $\pm{0.008}$ & 90.8 & 96.8 & 93.7 $\pm{0.8}$\\
 & No token coords & BIO & 0.037 $\pm{0.004}$ & 96.0  & 98.2  & 97.1 $\pm{0.9}$\\
 & No GloVe & BIO & 0.039 $\pm{0.002}$ & 96.0 & 98.6 & 97.3 $\pm{0.4}$ \\
\midrule
Ours & All features & BI & 0.031 $\pm{0.004}$ & 95.5 & 99.0  & 97.2 $\pm{1.2}$\\
 & ELMo not fine-tuned & BI & 0.050 $\pm{0.006}$ & 91.6  & 98.6  & 94.9 $\pm{0.7}$\\
 & No ELMo & BI & 0.072 $\pm{0.010}$ & 92.2 & 97.2 & 94.6 $\pm{1.9}$\\
 & No token coords & BI & \textbf{0.029} $\pm{0.003}$ & 94.9 & \textbf{99.1} & 97.0 $\pm{1.1}$\\
 & No GloVe & BI & 0.033 $\pm{0.002}$ & 95.9 & 99.0 & 97.4 $\pm{0.6}$ \\
\midrule
Koshorek et al. & & BI & 0.266 $\pm{0.004}$ & 20.0 & 96.0 & 33.0 $\pm{0.2}$ \\
\bottomrule
\end{tabular}
\label{table:main-table}
\end{table*}

% Booktabs table:
\begin{table}[b!]
\centering
\caption{Detailed task-based evaluation of our model and the model proposed in \cite{koshorek-etal-2018-text} trained and tested on marriage announcements. Scores shown are the average of three experiments. Standard deviations are given for F1.}
\begin{tabular}{@{}llrrr@{}}
\toprule
Model & Entity Type & Precision & Recall & F1\\ \midrule
Ours (BIO) & Bride & 97.8 &	98.9 &	98.4 $\pm{0.2}$  \\
With pos. vectors &  Groom & 97.6 &	98.5 &	98.1 $\pm{0.2}$ \\
&  BrideResidence & 97.7 &	98.8 &	98.3 $\pm{0.2}$ \\
&  GroomResidence & 97.6 &	99.2 &	98.4 $\pm{0.3}$ \\
&  WeddingDate & 92.8 &	95.0 &	93.9 $\pm{0.9}$ \\
\midrule
Ours (BIO) & Bride & 95.5 &	98.8 &	97.1 $\pm{1.5}$  \\
No pos. vectors &  Groom & 95.5 &	98.2 &	96.9 $\pm{1.4}$ \\
&  BrideResidence & 97.6 &	98.6 &	98.1 $\pm{0.3}$ \\
&  GroomResidence & 97.5 &	99.2 &	98.3 $\pm{0.3}$ \\
&  WeddingDate & 71.4 &	89.8 &	77.1 $\pm{21}$ \\
\midrule
Ours (BI) & Bride	& 96.0 &	99.3 &	97.6 $\pm{1.0}$ \\
With pos. vectors & Groom &	95.9 &	98.8 &	97.3 $\pm{1.2}$ \\
& BrideResidence &	96.9 & 98.9 &	97.9 $\pm{0.5}$ \\
& GroomResidence	& 97.0 &	99.3 & 98.1 $\pm{0.7}$ \\
& WeddingDate &	76.3	& 93.4 &	83.7 $\pm{6.7}$\\
\midrule
Koshorek et al. & Bride	& 15.1 &	97.6 &	26.2 $\pm{0.1}$ \\
& Groom &	15.1 &	95.2 &	26.0 $\pm{0.1}$\\
& BrideResidence &	28.8 & 93.3 &	44.0 $\pm{0.1}$  \\
& GroomResidence	& 29.7 &	96.7 & 45.4 $\pm{0.1}$ \\
& WeddingDate &	34.3	& 94.3 &	50.3 $\pm{1.1}$ \\
\bottomrule
\end{tabular}

\label{table:task-detail}
\end{table}

\noindent \textbf{Task-based evaluation} \quad As noted in \cite{Manning_rethinkingtext}, standard segmentation evaluation metrics fail to account for the fact that some errors are worse than others. For example, in Fig. \ref{fig:seg_examples}, the phrase \textit{(No cards.)} could have been excluded from the first segment without any loss of critical information. On the other hand, the segmentation error shown in Fig. \ref{fig:example_with_coords}A is serious, as it places the Bride and Groom in different segments.

As an alternative way of measuring segmentation accuracy, we developed a task-based evaluation method. For all of the marriage announcements in our test set, a set of marriage-related entities was hand-labeled (\texttt{Bride}, \texttt{Groom}, \texttt{WeddingDate}, etc.) We use these entities in the task-based evaluation as follows. We iterate over the ground truth segments, finding the predicted segment with the most overlap, as measured by the number of characters shared. We then compare the entities found in the predicted segment with entities found in the matching ground truth segment. We define true positives as entities found in both the predicted and ground truth segments; false positives as entities found in the predicted segment, but not in the ground truth segment; and false negatives as entities found in the ground truth segment, but not in the predicted segment. We use these counts of true positives, false positives, and false negatives to compute precision, recall, and F1 scores.

\medskip
For each experiment, we report the results of three trials in which we trained on the combined training and dev sets and evaluated performance on the test set. 

\section{Results}
Table \ref{table:main-table} shows the final results of experiments with our model and the model of \cite{koshorek-etal-2018-text}. We choose to compare our model's performance against that of \cite{koshorek-etal-2018-text}, rather than other recent topic segmentation models, because of the availability of the code used in \cite{koshorek-etal-2018-text}. We tuned the hyperparameters of the latter model by training on our training set and calculating \(P_k\) on the dev set for different hyperparameter values. Final reported results used the optimal hyperparameters found via this procedure and used the combined training and dev sets for training and the test set for evaluation.

As noted above, we used three token labels (\texttt{B-Marriage}, \texttt{I-Marriage}, and \texttt{O}), because certain sections of text are not part of a particular marriage announcement. This is distinct from most topic segmentation approaches, in which all parts of the document are assumed to belong to a segment, and the task is formulated as finding either the beginning of each segment or the end of each segment. We refer to this as a BI tagging scheme by analogy to the BIO tagging scheme common in sequence-tagging tasks; the model of \cite{koshorek-etal-2018-text} also uses such an approach.\footnote{Technically, the model of \cite{koshorek-etal-2018-text} tags each sentence as either the end of a segment or not the end of a segment. This is not a BI scheme since it predicts the ends of segments rather than the beginnings. Nonetheless, it is conceptually equivalent; the key point is that there are only two class labels being predicted.} In order to make a more direct comparison of our method to the method of \cite{koshorek-etal-2018-text}, we also made a BI-tagged version of our data, in which the labels for any \texttt{O}-tagged tokens were converted to \texttt{B-Marriage} or \texttt{I-Marriage}.

Our model significantly outperforms that of \cite{koshorek-etal-2018-text}, as measured both by \(P_k\) and by the task-based evaluation method. The model of \cite{koshorek-etal-2018-text} shows particularly low precision in the task-based evaluation, indicating that it tends to under-segment. This is consistent with our observation that the Punkt sentence tokenizer used by \cite{koshorek-etal-2018-text} tends to group multiple marriage announcements together as one segment. Across our entire data set, 48\% of all segment boundaries did not align with sentence boundaries identified by the Punkt sentence tokenizer. 

Table \ref{table:main-table} also shows experiments to determine the contribution of the ELMo embeddings, token positions, GloVe embeddings, and BIO encoding to the performance of our model. The best performance, as measured by the highest F1 score in the task-based evaluation, was obtained when ELMo embeddings, GloVe embeddings, and token position vectors were included as features, and when \texttt{B-Marriage}, \texttt{I-Marriage}, and \texttt{O} were used as token labels.

When using BIO tags, use of ELMo embeddings increased the F1 score by 4\% (from 93.7\% to 97.7\%). A significant part of this increase can be attributed to fine-tuning the ELMo language model on text from the same domain. Without fine-tuning, the F1 score was 95.5\%, while with fine-tuning it was 97.7\%. When using BI tags, the use of our fine-tuned ELMo model increased the F1 score by 2.6\% (from 94.6\% to 97.2\%). Fine-tuning ELMo increased the F1 score by 2.3\% (from 94.9\% to 97.2\%).

\begin{figure*}
    \centering
    \includegraphics[]{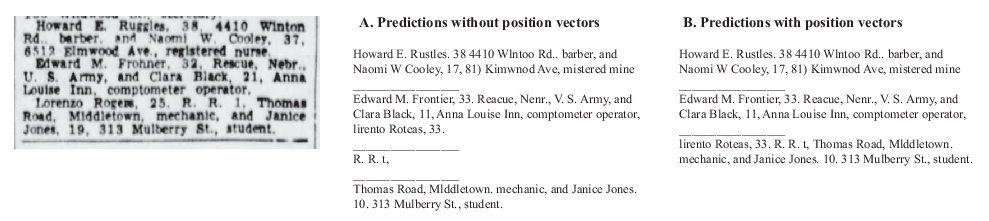}
    \caption{Example of a marriage announcement that was correctly segmented when token position vectors were used as a feature, but incorrectly segmented otherwise. The predictions from models trained with or without token position vectors are shown for the same article used in Fig. \ref{fig:sentence_tokenization}.}
    \label{fig:example_with_coords}
\end{figure*}

The contribution of token position vectors was smaller, boosting the F1 score from 97.1\% without position vectors to 97.7\% with position vectors when using BIO tags. When using BI tags, the use of token position vectors increased the F1 score from 97.0\% to 97.2\%. Table \ref{table:task-detail} shows greater detail for the task-based evaluation. A small boost in precision and F1 score is seen for all of the individual entity types we tested when position vectors are used (Table \ref{table:task-detail}). However, in all cases, the size of the effect is small and does not achieve statistical significance in these experiments (\(p = 0.32\) for overall task-based F1 score with BIO tags, Student's t-test). This lack of statistical significance might reflect the small effect size and limited number of experiments. Positional information appears to improve segmentation of only a small subset of articles.

Fig. \ref{fig:example_with_coords} shows an example of an article whose segmentation was improved by the use of token position vectors. Without token position vectors, the last marriage announcement in the article was both under- and over-segmented; the groom in the last segment was erroneously included in the previous announcement, while part of his address and the bride were split into separate segments. With position vectors, the article was correctly segmented. These different outcomes were seen in all three replicates of each experiment. The inclusion of the groom into the previous announcement may have occurred because of OCR errors that converted the preceding period into a comma and converted the first letter of the groom’s name from uppercase to lowercase. Even without these cues to start a new segment, the model trained with token position vectors still correctly predicted the beginning of the segment. 

Our best-performing model uses fixed GloVe embeddings in addition to contextual ELMo embeddings, but ablation experiments (Table \ref{table:main-table}) show that the contribution of GloVe embeddings is small, and the effect is not statistically significant (\(p = 0.23\) for task-based F1 score with BIO tags, Student's t-test). At best, the GloVe embeddings make only a small contribution to model performance. We speculate that ELMo embeddings contribute essentially the same information, in a more sophisticated form, that GloVe embeddings contribute, therefore rendering the GloVe embeddings nearly superfluous.

Using three token labels (\texttt{B-Marriage}, \texttt{I-Marriage}, and \texttt{O}) rather than two, also has only a minor impact on performance. When \(P_k\) is used as the evaluation metric, the BI-tagging scheme appears to perform better than the BIO-tagging scheme, an effect which is on the border of statistical significance (\(p = 0.05\) for models using all features, Student's t-test). It is perhaps not surprising that performance as measured by \(P_k\) is enhanced by using the BI-tagging scheme. As noted above, in order to calculate \(P_k\) for BIO-tagged predictions, we had to convert them to a BI-tagging scheme, since the \(P_k\) calculation assumes every token is part of a segment. No such conversion was necessary when the model was trained on BI-tagged labels. Thus, when we trained on BI-tagged data, the training objective matched the \(P_k\) evaluation metric more closely than when we trained on BIO-tagged data. 

We believe the task-based evaluation is a more meaningful metric for our purposes, since it quantifies errors that would directly affect information extraction. By this metric, we see that the BIO-tagging scheme achieves slightly higher precision and a higher F1 score than the BI-tagging scheme, although these improvements are not statistically significant (\(p = 0.36\) for precision and \(p = 0.5\) for F1, Student's t-test). The performance for \texttt{WeddingDate}, a fact typically located outside of marriage announcement segments, shows a notable increase when \texttt{O} labels are used in addition to \texttt{B-Marriage} and \texttt{I-Marriage}, though this effect is on the border of statistical significance (\(p = 0.06\) for the task-based F1 score, Student's t-test). We speculate that the inclusion of the \texttt{O} label allows the model to specialize more in learning specific features associated with non-marriage text, and therefore allows the model to do a better job of excluding those sections from adjacent segments.

\section{Conclusions}
Here we have presented a novel neural architecture for segmenting image-derived text of newspaper marriage announcements. Our use case differs from most previous contexts in which topic segmentation has been employed due to the messiness of our text, the lack of narrative structure in our data, topical similarity between segments, and the hierarchical structure of our data. While our model draws inspiration from recent supervised approaches to topic segmentation, we also deviate from most recent proposals in several respects. In particular, we perform segmentation on the token level, rather than at a higher level such as the sentence level, we utilize a pre-trained language model trained on in-domain text, and we experiment with including spatial information for each token.

Our results demonstrate the importance of performing segmentation at the token level, rather than the sentence level. This is particularly apparent when comparing the performance of our architecture against the performance of the architecture proposed in \cite{koshorek-etal-2018-text}. Fine-tuning our ELMo model on in-domain text also contributes significantly to the performance of our model. We speculate that this improvement is in part attributable to the fact that this fine-tuning allows the ELMo model to generate sensible contextual token embeddings for words containing common OCR errors, although our results are compatible with other explanations. For example, this fine-tuning may simply allow the ELMo model to generate contextual embeddings that better reflect the domain of historical newspaper text, irrespective of any OCR errors.  

The contribution of spatial features is less clear. We saw a small increase in precision and F1 score, as measured by the task-based evaluation method, when position vectors were used, particularly for an entity type typically located outside or at the boundaries of segments (\texttt{WeddingDate}). However, the performance improvement did not achieve statistical significance in our experiments. Further studies are needed to help clarify the conditions under which spatial features contribute significantly to model performance. Different types of data, such as invoices containing many line items, may benefit more from spatial features than do marriage announcements.

We use a BIO-tagging scheme in order to capture hierarchy in our data: text not describing a specific couple is labeled \texttt{O} and is not considered to be part of a \texttt{Marriage} segment. This is necessary in our case because of the way information is structured in articles in our data set, but we show that our architecture also performs well when simple linear segmentation is the goal. In our production system, we assume that any facts, such as \texttt{WeddingDate}, located outside of \texttt{Marriage} segments apply to all subsequent segments in the same document until another \texttt{WeddingDate} is encountered. The BIO-tagging scheme could be extended to capture more sophisticated hierarchies, for example by including labels such as \texttt{GlobalInformation}, \texttt{Subheading}, etc. 

Our work suggests numerous extensions to similar problems. This tagging scheme could easily be extended to classify multiple types of segments. For example, in segmenting different articles in a news feed, multiple tags such as B-Sports, I-Sports, B-Politics, I-Politics, etc., could be used to simultaneously segment and categorize the articles. While here we use an ELMo model to generate contextual token embeddings, other types of models, such as BERT, could be used for similar purposes \cite{devlin2018bert}. One outstanding issue is how BERT's WordPiece tokenization would handle OCR errors commonly encountered in our data set. 

The methods described here could also be applied to other sources of noisy text, such as speech-to-text transcripts or text derived from handwriting recognition. They are especially well-suited to segmenting text lacking complete sentences or clear sentence boundaries. Such applications could include segmentation of receipts, invoices, recipes, song lyrics, or movie subtitles.

\balance
\bibliographystyle{IEEEtran} 
\bibliography{list_seg_refs}

% Generated by IEEEtran.bst, version: 1.14 (2015/08/26)
\begin{thebibliography}{10}
\providecommand{\url}[1]{#1}
\csname url@samestyle\endcsname
\providecommand{\newblock}{\relax}
\providecommand{\bibinfo}[2]{#2}
\providecommand{\BIBentrySTDinterwordspacing}{\spaceskip=0pt\relax}
\providecommand{\BIBentryALTinterwordstretchfactor}{4}
\providecommand{\BIBentryALTinterwordspacing}{\spaceskip=\fontdimen2\font plus
\BIBentryALTinterwordstretchfactor\fontdimen3\font minus
  \fontdimen4\font\relax}
\providecommand{\BIBforeignlanguage}[2]{{%
\expandafter\ifx\csname l@#1\endcsname\relax
\typeout{** WARNING: IEEEtran.bst: No hyphenation pattern has been}%
\typeout{** loaded for the language `#1'. Using the pattern for}%
\typeout{** the default language instead.}%
\else
\language=\csname l@#1\endcsname
\fi
#2}}
\providecommand{\BIBdecl}{\relax}
\BIBdecl

\bibitem{beeferman1999statistical}
D.~Beeferman, A.~Berger, and J.~Lafferty, ``Statistical models for text
  segmentation,'' \emph{Machine Learning}, vol.~34, no. 1-3, pp. 177--210,
  1999.

\bibitem{webster-1992}
J.~J. Webster and C.~Kit, ``Tokenization as the initial phase in {NLP},'' in
  \emph{Proc. of the 14th Conf. on Comput. Linguist.}\hskip 1em plus 0.5em
  minus 0.4em\relax USA: Assoc. for Comput. Linguist., 1992, p. 1106–1110.

\bibitem{hearst-1997}
M.~A. Hearst, ``{TextTiling}: Segmenting text into multi-paragraph subtopic
  passages,'' \emph{Comput. Linguist.}, vol.~23, no.~1, p. 33–64, Mar. 1997.

\bibitem{riedl2012topictiling}
M.~Riedl and C.~Biemann, ``{TopicTiling}: a text segmentation algorithm based
  on {LDA},'' in \emph{Proce. of ACL 2012 Student Research Workshop}.\hskip 1em
  plus 0.5em minus 0.4em\relax Assoc. for Comput. Linguist., 2012, pp. 37--42.

\bibitem{koshorek-etal-2018-text}
O.~Koshorek, A.~Cohen, N.~Mor, M.~Rotman, and J.~Berant, ``Text segmentation as
  a supervised learning task,'' in \emph{Proc. of the 2018 Conf. of the North
  {A}merican Chapter of the Assoc. for Comput. Linguist}.\hskip 1em plus 0.5em
  minus 0.4em\relax New Orleans, USA: Assoc. for Comput. Linguist., Jun. 2018,
  pp. 469--473.

\bibitem{chen2009global}
H.~Chen, S.~Branavan, R.~Barzilay, and D.~R. Karger, ``Global models of
  document structure using latent permutations,'' in \emph{Proc. the 2009
  Annual Conf. of the North American Chapter of the Assoc. for Computat.
  Linguist.}\hskip 1em plus 0.5em minus 0.4em\relax Assoc. for Comput.
  Linguist., 2009, p. 371–379.

\bibitem{glavavs2020two}
G.~Glava{\v{s}} and S.~Somasundaran, ``Two-level transformer and auxiliary
  coherence modeling for improved text segmentation,'' \emph{arXiv preprint
  arXiv:2001.00891}, 2020.

\bibitem{AzziTheFS}
A.~A. Azzi, H.~Bouamor, and S.~Ferradans, ``The {FinSBD}-2019 shared task :
  Sentence boundary detection in {PDF} noisy text in the financial domain,'' in
  \emph{Proc. of the First Workshop on Financial Technology and Natural
  Language Processing (FinNLP@IJCAI 2019)}, 2019, pp. 74--80.

\bibitem{sheikh2017topic}
I.~Sheikh, D.~Fohr, and I.~Illina, ``Topic segmentation in asr transcripts
  using bidirectional rnns for change detection,'' in \emph{2017 IEEE Automatic
  Speech Recognition and Understanding Workshop (ASRU)}.\hskip 1em plus 0.5em
  minus 0.4em\relax IEEE, 2017, pp. 512--518.

\bibitem{peters-etal-2018-deep}
\BIBentryALTinterwordspacing
M.~Peters, M.~Neumann, M.~Iyyer, M.~Gardner, C.~Clark, K.~Lee, and
  L.~Zettlemoyer, ``Deep contextualized word representations,'' in \emph{Proc.
  of the 2018 Conf. of the North {A}merican Chapter of the Assoc. for Comput.
  Linguist.}\hskip 1em plus 0.5em minus 0.4em\relax New Orleans, Louisiana:
  Assoc. for Comput. Linguist., Jun. 2018, pp. 2227--2237. [Online]. Available:
  \url{https://www.aclweb.org/anthology/N18-1202}
\BIBentrySTDinterwordspacing

\bibitem{hearst1994multi}
M.~A. Hearst, ``Multi-paragraph segmentation of expository text,'' in
  \emph{Proceedings of the 32nd Annual Meeting of the Assoc. for Comput.
  Linguist.}\hskip 1em plus 0.5em minus 0.4em\relax Assoc. for Comput.
  Linguist., 1994, pp. 9--16.

\bibitem{choi2000advances}
F.~Y.~Y. Choi, ``Advances in domain independent linear text segmentation,'' in
  \emph{1st Meeting of the North {A}merican Chapter of the Assoc. for Comput.
  Linguist.}, 2000.

\bibitem{blei2003latent}
D.~M. Blei, A.~Y. Ng, and M.~I. Jordan, ``Latent {Dirichlet} allocation,''
  \emph{Journal of Machine Learning Research}, vol.~3, no. Jan, pp. 993--1022,
  2003.

\bibitem{mota-etal-2019-beamseg}
P.~Mota, M.~Eskenazi, and L.~Coheur, ``{B}eam{S}eg: A joint model for
  multi-document segmentation and topic identification,'' in \emph{Proc. of the
  23rd Conf. on Comput. Natural Language Learning (CoNLL)}.\hskip 1em plus
  0.5em minus 0.4em\relax Hong Kong, China: Assoc. for Comput. Linguist., Nov.
  2019, pp. 582--592.

\bibitem{glavavs2016unsupervised}
G.~Glava{\v{s}}, F.~Nanni, and S.~P. Ponzetto, ``Unsupervised text segmentation
  using semantic relatedness graphs,'' in \emph{Proc. of the Fifth Joint Conf.
  on Lexical and Computat. Semantics}.\hskip 1em plus 0.5em minus 0.4em\relax
  Assoc. for Comput. Linguist., 2016, pp. 125--130.

\bibitem{mikolov2013efficient}
T.~Mikolov, K.~Chen, G.~Corrado, and J.~Dean, ``Efficient estimation of word
  representations in vector space,'' in \emph{Proc. of Workshop at ICLR}, 2013.

\bibitem{Schuster1997}
M.~Schuster and K.~Paliwal, ``Bidirectional recurrent neural networks,''
  \emph{Signal Processing, IEEE Transactions on}, vol.~45, pp. 2673 -- 2681, 12
  1997.

\bibitem{hochreiter1997long}
S.~Hochreiter and J.~Schmidhuber, ``Long short-term memory,'' \emph{Neural
  Computation}, vol.~9, no.~8, pp. 1735--1780, 1997.

\bibitem{arnold_et_al_2019}
S.~Arnold, R.~Schneider, P.~Cudré-Mauroux, F.~A. Gers, and A.~Löser,
  ``{SECTOR}: A neural model for coherent topic segmentation and
  classification,'' \emph{Transactions of the Assoc. for Comput. Linguist.},
  vol.~7, pp. 169--184, 2019.

\bibitem{chang2019language}
M.-W. Chang, K.~Toutanova, K.~Lee, and J.~Devlin, ``Language model pre-training
  for hierarchical document representations,'' \emph{arXiv preprint
  arXiv:1901.09128}, 2019.

\bibitem{vaswani2017attention}
A.~Vaswani, N.~Shazeer, N.~Parmar, J.~Uszkoreit, L.~Jones, A.~N. Gomez,
  {\L}.~Kaiser, and I.~Polosukhin, ``Attention is all you need,'' in
  \emph{Advances in Neural Information Processing Systems}, 2017, pp.
  5998--6008.

\bibitem{Manning_rethinkingtext}
C.~D. Manning, ``Rethinking text segmentation models: An information extraction
  case study,'' University of Sydney, Tech. Rep., 1998.

\bibitem{DBLP:journals/corr/cmp-lg-9505040}
L.~A. Ramshaw and M.~P. Marcus, ``Text chunking using transformation-based
  learning,'' in \emph{Third Workshop on Very Large Corpora}, 1995, pp. 82--94.

\bibitem{lample-etal-2016-neural}
G.~Lample, M.~Ballesteros, S.~Subramanian, K.~Kawakami, and C.~Dyer, ``Neural
  architectures for named entity recognition,'' in \emph{Proc. of the 2016
  Conf. of the North {A}merican Chapter of the Assoc. for Computat.ƒ
  Linguist.}\hskip 1em plus 0.5em minus 0.4em\relax San Diego, California:
  Assoc. for Comput. Linguist., Jun. 2016, pp. 260--270.

\bibitem{ma-hovy-2016-end}
X.~Ma and E.~Hovy, ``End-to-end sequence labeling via bi-directional
  {LSTM}-{CNN}s-{CRF},'' in \emph{Proc. of the 54th Annual Meeting of the
  Assoc. for Comput. Linguist.}\hskip 1em plus 0.5em minus 0.4em\relax Berlin,
  Germany: Assoc. for Comput. Linguist., Aug. 2016, pp. 1064--1074.

\bibitem{pennington-etal-2014-glove}
J.~Pennington, R.~Socher, and C.~Manning, ``{GloVe}: Global vectors for word
  representation,'' in \emph{Proc. of the 2014 Conf. on Empirical Methods in
  Natural Language Processing ({EMNLP})}.\hskip 1em plus 0.5em minus
  0.4em\relax Doha, Qatar: Assoc. for Comput. Linguist., Oct. 2014, pp.
  1532--1543.

\bibitem{JMLR:v12:collobert11a}
R.~Collobert, J.~Weston, L.~Bottou, M.~Karlen, K.~Kavukcuoglu, and P.~Kuksa,
  ``Natural language processing (almost) from scratch,'' \emph{Journal of
  Machine Learning Research}, vol.~12, no.~76, pp. 2493--2537, 2011.

\bibitem{Dozat2016IncorporatingNM}
T.~Dozat, ``Incorporating {Nesterov} momentum into {Adam},'' in \emph{ICLR
  Workshop}, 2016, pp. (1):2013--2016.

\bibitem{fournier-2013-evaluating}
C.~Fournier, ``Evaluating text segmentation using boundary edit distance,'' in
  \emph{Proc. of the 51st Annual Meeting of the Assoc. for Comput.
  Linguist.}\hskip 1em plus 0.5em minus 0.4em\relax Sofia, Bulgaria: Assoc. for
  Comput. Linguist., Aug. 2013, pp. 1702--1712.

\bibitem{devlin2018bert}
J.~Devlin, M.-W. Chang, K.~Lee, and K.~Toutanova, ``{BERT}: Pre-training of
  deep bidirectional transformers for language understanding,'' in \emph{Proc.
  of the 2019 Conf. of the North {A}merican Chapter of the Assoc. for Computat.
  Linguist.}\hskip 1em plus 0.5em minus 0.4em\relax Assoc. for Comput.
  Linguist., 2019, pp. 4171--4186.

\end{thebibliography}
\end{document}